\documentclass[11pt]{article}
\usepackage{acl2015}
\usepackage{times}
\usepackage{url}
\usepackage{latexsym}
\usepackage{graphicx}
\usepackage{booktabs}
\usepackage{color}
\usepackage{enumitem}

\title{Talking to the crowd: What do people react to in online discussions?}

\author{Aaron Jaech, Vicky Zayats, Hao Fang, Mari Ostendorf and Hannaneh Hajishirzi\\
  Dept. of Electrical Engineering \\
  University of Washington\\
  {\tt \{ajaech,vzayats,hfang,ostendor,hannaneh\}@uw.edu} }

\date{}

\begin{document}
\maketitle
\begin{abstract}
This paper addresses the question of how language use affects community reaction
to comments in online discussion forums, and the relative importance of the
message vs.\ the messenger. A new comment ranking task is proposed based on
community annotated karma in Reddit discussions, which controls for topic and
timing of comments. Experimental work with discussion threads from six
subreddits shows that the importance of different types of language features
varies with the community of interest.
\end{abstract}

\section{Introduction}
\label{sec:intro}
Online discussion forums are a popular platform for people to share their views about current events and learn about issues of concern to them.  Discussion forums tend to specialize on different topics, and people participating in them form communities of interest.  The reaction of people within a community to comments posted provides an indication of community endorsement of opinions and value of information. In most discussions, the vast majority of comments spawn little reaction.  In this paper, we look at whether (and how) language use affects the reaction, compared to the relative importance of the author and timing of the post.

Early work on factors that appear to influence crowd-based judgments of comments in the Slashdot forum \cite{LampeRes04} indicate that timing, starting score, length of the comment, and poster anonymity/reputation appear to play a role (where anonymity has a negative effect). Judging by differences in popularity of various discussion forums, topic is clearly important.
Evidence that language use also matters is provided by recent work
\cite{danescu2012you,lakkaraju2013s,althoff2014ask,tan2014effect}. Teasing these different factors apart, however, is a challenge. The work presented in this paper provides additional insight into this question by controlling for these factors in a different way than previous work and by examining multiple communities of interest. Specifically, using data from Reddit discussion forums, we look at the role of author reputation as measured in terms of a karma k-index, and control for topic and timing by ranking comments in a constrained window within a discussion.

The primary contributions of this work include findings about the role of author reputation and variation across communities in terms of aspects of language use that matter, as well as the problem formulation, associated data collection, and development of a variety of features for characterizing informativeness, community response, relevance and mood.

\section{Data}
\label{sec:data}
Reddit\footnote{http://www.reddit.com} 
is the largest public
online discussion forum with a wide variety of subreddits, which makes it a
good data source for studying how textual content in a discussion impacts the response of the crowd.
On Reddit, people initiate a discussion thread with a post (a question, a link to a news item, etc.), and others respond with comments.
Registered users vote on which posts and comments are important. 
The total amount of up votes minus the down votes (roughly) is called {\it karma}; it provides an indication of community endorsement and popularity of a comment, as used in \cite{lakkaraju2013s}.
Karma is valued as it impacts the order in which the posts or comments are displayed, with 
the high karma content rising to the top.
Karma points are also accumulated by members of the discussion forum as a function of the karma associated with their comments.

The Reddit data is highly skewed. Although there are thousands of active communities,
only a handful of them are large. Similarly, out of the more than a million comments
made per day\footnote{http://www.redditblog.com/2014/12/reddit-in-2014.html}, most of
them receive little to no attention; 
the distributions of positive comment karma and author karma are Zipfian.
Slightly more than half of all comments have exactly
one karma point (no votes beyond the author), and only
5\% of comments have less than one karma point. 



\begin{table}[t]
	\centering
{\small
	\begin{tabular}{|lcc|}
		\toprule
		{\bf subreddit}  \
		& {\bf \# Posts} & {\bf \# Comments/Post} \\
		\hline\rule[-1ex]{0pt}{3.5ex}%
		{\sc fitness} 		& 3K 	& 16.3 	\\
		{\sc askscience} 	& 4K	& 8.8 	\\
		{\sc politics} 		& 7K 	& 23.7 	\\
		{\sc askwomen} 		& 4K 	& 50.5 	\\
		{\sc askmen} 			& 4K 	& 58.3 	\\
		{\sc worldnews} 	& 12K & 26.1 	\\
		\bottomrule
	\end{tabular}
}
	\caption{Data collection statistics.}
	\label{tab:data}
\end{table}


For this study, we downloaded all the posts and associated comments made to six subreddits over a few weeks, as summarized in Table~\ref{tab:data}, as well as karma of participants in the discussion\footnote{Our data collection is available online at https://ssli.ee.washington.edu/tial/data/reddit}.
All available comments on each post were downloaded at least 48 hours after the post was
made.\footnote{Based on our initial look at the data, we noticed that most posts
receive all of their comments within 48 hours. 
Some comments are deleted before we are able to download them.} 


\section{Uptake Factors}
\label{sec:factors}
\label{factors_sec}
Factors other than the language use that influence whether a comment will have uptake from the community include the topic, the timing of the message, and the messenger. These factors are all evident in the Reddit discussions.  
Some subreddits are more popular and thus have higher karma comments than
others, reflecting the influence of topic.  Comments that are posted early in
the discussion are more likely to have high karma, since they have more potential responses.  

Previous studies on Twitter show that the reputation of the author substantially increases the chances of the retweet
\cite{suh2010want,cha2010measuring}, 
and reputation is also raised as a factor in Slashdot \cite{LampeRes04}.
On Reddit most users are anonymous, but it is possible that members of a forum
become familiar with particular usernames associated with high karma comments.
In order to see how important personal reputation is, we looked at how often the
top karma comments are associated with the top karma participants in the
discussion.  Since an individual's karma can be skewed by a few very popular
posts, we measure reputation instead using a measure we call the {\it k-index},
defined to be equal to the number of comments in each user's history
that have karma $\geq k$. The k-index is
analgous to the h-index \cite{hirsch2005index} and arguably a better indicator of extended impact than total karma. 

The results in Table~\ref{tab:hindex_results} address the question of whether
the top karma comments always come from the top karma person. The Top1 column shows the percentage of threads where
the top karma comment in a discussion happens to be made by the highest k-index person participating in the discussion; the next column shows the percentage of threads where the comment comes from any one of the top 3 k-index people. We find that, in fact, the highest karma comment in a discussion is rarely from the highest k-index people. The highest percentage is in {\sc askscience}, where expertise is more highly valued. If we consider whether any one of the multiple comments that the top k-index person made is the top karma comment in the discussion, then the frequency is even lower.

%

\begin{table}
	\centering
{\small
	\begin{tabular}{|lcc|}
		\toprule
		& {\bf Top1}
		& {\bf Top3} 
		\\
		\hline\rule[-1ex]{0pt}{3.5ex}%
		{\sc askscience} & 9.3 & 25.9   \\
		{\sc fitness} 	 & 1.4 & 12.3   \\
		{\sc politics} 	 & 0.3 & 7.4    \\
		{\sc askwomen}   & 2.2 & 13.5   \\
		{\sc askmen }	 & 3.9 & 11.9   \\
		{\sc worldnews}  & 3.1 & 6.4    \\
		\bottomrule
	\end{tabular}
}
	\caption{Percentage of discussions where the top comment is made by the top k-index person (or top 3 people) in the discussion.}
	\label{tab:hindex_results}
\end{table}

%

\section{Methods}
\label{sec:methods}

\subsection{Tasks}

Having shown that the reputation of the author of a post is not a dominating factor in predicting high karma comments, we propose to control for topic and timing by ranking a set of 10 comments that were made consecutively in a short window of
time within one discussion thread according to the karma they finally received. The ranking has access to the comment history about these posts.  This simulates the view of an early reader of these posts, i.e., without influence of the ratings of others, so that the language content of the post is more likely to have an impact.
Very long threads are sampled, so that these do not dominate the set of lists.
Approximately 75\% of the comment lists are designated for training and the rest is for
testing, with splits at the discussion thread level. 
Here, feature selection is based on mean precision of the top-ranked comment (P@1), so as to emphasize learning the rare high karma events. 
(Note that P@1 is equivalent to accuracy but allows for any top-ranking comment to count as correct in the case of ties.)  The system performance is evaluated using both P@1 and normalized discounted cumulative gain (NDCG) \cite{NDCG05}, which is a standard criterion for ranking evaluation when the samples to be ranked have meaningful differences in scores, as is the case for karma of the comments.  



In addition, for analysis purposes, we report results for three surrogate tasks that can be used in the ranking problem: i) the binary ranker trained on 
all comment pairs within each list,
in which low karma comments dominate, ii) a positive vs.\ negative karma classifier, and iii) a high vs.\ medium karma classifier.  All use class-balanced data; the second two are trained and tested on a biased sampling of the data, where the pairs need not be from the same discussion thread.

\subsection{Classifier}
We use the support vector machine (SVM) rank algorithm
\cite{joachims2002optimizing} to predict a rank order for each list of comments.
The SVM is trained to predict which of a pair of
comments has higher karma. 
The error term penalty parameter is tuned to maximize P@1 on a held-out
validation set (20\% of the training samples).

Since much of the data includes low-karma comments, there will be a tendancy for the learning to emphasize features that discriminate comments at the lower end of the scale. In order to learn features that improve P@1, and to understand the relative importance of different features, we use a greedy 
automatic feature selection process that incrementally adds one
feature whose resulting feature set achives the highest P@1 on the validation
set.
Once all features have been used, we select the model with the subset of features
that obtains the best P@1 on the validation set.


\subsection{Features}
\label{sec:features}
The features are designed to capture several key attributes that we hypothesize
are predictive of comment karma motivated by related work. 
The features are categorized in groups as summarized below, with details in
supplementary material.
\begin{itemize}[noitemsep, topsep=0pt, leftmargin=*]
\item {\bf Graph and Timing (G\&T):} A baseline that captures discourse history (response structure) and comment timing, but no text content.
\item {\bf Authority and Reputation (A\&R):} K-index, whether the commenter was the original poster, and in some subreddits ``flair'' (display next to a
comment author's username that is subject to a cursory verification by moderators).
\item {\bf Informativeness (Info.):} Different indicators suggestive of informative content and novelty, including various word counts, named entity counts, urls, and unseen n-grams.
\item {\bf Lexical Unigrams (Lex.):} Miscellaneous word class indicators, puncutation, and part-of-speech counts
\item {\bf Predicted Community Response (Resp.):} Probability scores from surrogate classification tasks (reply vs.\ no reply, positive vs.\ negative sentiment)
to measure the community response of a comment using bag-of-words predictors. 
\item {\bf Relevance (Rel.):} Comment similarity to the parent, post and title in terms of topic, computed with three methods: i) a distributed vector representation of topic using a non-negative matrix factorization
(NMF) model \cite{xu2003document}, ii) the average of skip-gram word embeddings
\cite{mikolov2013distributed}, and iii) word set Jaccard similarity \cite{strehl2000impact}. 
\item {\bf Mood:} Mean and std.\ deviation of sentence sentiment in the comment; word list indicators for politeness, argumentativeness and profanity.
\item {\bf Community Style (Comm.):} Posterior probability of each subreddit given the comment using a bag-of-words model. 
\end{itemize}




The various word lists are motivated by feature exploration studies in surrogate tasks.  For example, projecting words to a two dimensional space of positive vs.\ negative and likelihood of reply showed that self-oriented pronouns were more likely to have no response and second-person pronouns were more likely to have a negative response. The
politeness and argumentativeness/profanity 
lists are generated by starting with hand-specified seed lists used to
train an SVM to classify word embeddings \cite{mikolov2013distributed} into 
these categories, and expanding the lists with 500 words farthest from the decision boundary. 

Both the NMF and the skip-gram topic models use a cosine distance to determine topic similarity, with 300 as the word embedding dimension. Both are trained on approximately 2 million comments in high karma posts taken across a wide variety of subreddits. We use topic models in various measures of comment relevance to the discussion, but we do not use topic of the comment on its own since topic is controlled for by ranking within a thread.

\section{Ranking Experiments}
\label{sec:ranking}
\begin{figure}[t]
	\centering
	\includegraphics[width=0.45\textwidth]{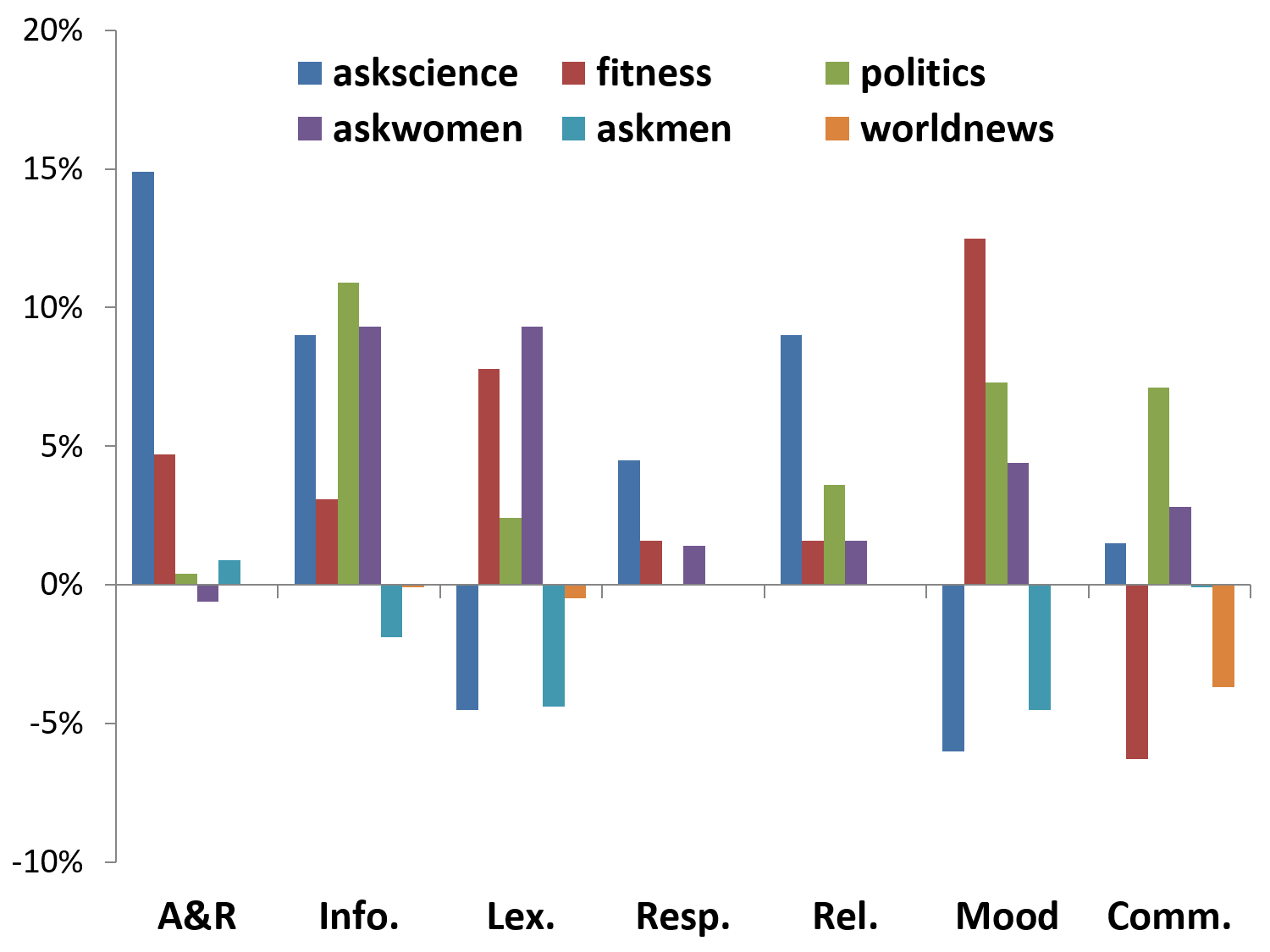}
	\caption{Relative improvement in P@1 over G\&T for individual feature groups.}
	\label{fig:barplot_p1}
\end{figure}

We present three sets of experiments on comment karma ranking, all of which show very different behavior for the different subreddits.
Fig.\,\ref{fig:barplot_p1} shows the relative gain in P@1 over the G\&T baseline associated with
using different feature groups.
The importance of the different features reflect the nature of the different communities.  The authority/reputation features help most for {\sc askscience}, consistent with our k-index study. Informativeness and relevance help all subreddits except {\sc askmen} and {\sc worldnews}. Lexical, mood and community style features are useful in some cases, but hurt others.  The predicted probability of a reply was least useful, possibly because of the low-karma training bias.

Tables\,\ref{tab:overallP} and \ref{tab:overallNDCG} summarize
the results for the P@1 and NDCG criteria using the greedy selection procedure (which optimizes P@1) compared to a random baseline and the G\&T baseline. The random baseline for P@1 is greater than 10\% because of ties. 
The G\&T baseline results show that the graph and timing features alone
obtain 21-32\% of top karma comments depending on subreddits.
Adding the textual features gives an improvement in P@1 performance over the G\&T baseline for all subreddits except {\sc askmen} and {\sc worldnews}. The trends for performance measured with NDCG are similar, but the benefit from textual features is smaller. The results in both tables show different ways of reporting performance of the same system, but the system has been optimized for P@1 in terms of feature selection.  In initial exploratory experiments, this seems to have a small impact: when optimizing for NDCG in feature selection we obtain 0.61 vs.\ 0.60 with the P@1-optimized features.

\begin{table}
	\centering
{\footnotesize
	\begin{tabular}{|l|ccc|}
		\toprule
		{\bf subreddit} & {\bf Random} & {\bf G\&T} & {\bf All} \\
		\hline\rule[-1ex]{0pt}{3.5ex}%
		{\sc askscience} 	& 15.9 & 21.8 & {\bf 25.3} \\
		{\sc fitness} 			& 19.4 & 22.1 & {\bf 27.3} \\
		{\sc politics} 		& 18.5 & 24.7 & {\bf 26.4} \\
		{\sc askwomen} 		& 17.6 & 24.9 & {\bf 28.0} \\
		{\sc askmen} 			& 18.2 & {\bf 31.4} & 29.1 \\
		{\sc worldnews} 		& 15.4 & {\bf 24.5} & 23.3 \\
		\hline\rule[-1ex]{0pt}{3.5ex}%
		Improvement 		& - &  42.9\% & 52.1\% \\
		\bottomrule
	\end{tabular}
}
	\caption{Test set precision of top one prediction (P@1) performance for specific subreddits.}
	\label{tab:overallP}
\end{table}

\begin{table}
	\centering
{\footnotesize
	\begin{tabular}{|l|ccc|}
		\toprule
		{\bf subreddit} & {\bf Random} & {\bf G\&T} & {\bf All} \\
		\hline\rule[-1ex]{0pt}{3.5ex}%
		{\sc askscience} 	& 0.53 & {\bf 0.60} & {\bf 0.60} \\
		{\sc fitness} 		& 0.57 & 0.61 & {\bf 0.62} \\
		{\sc politics} 		& 0.55 & 0.61 & {\bf 0.62} \\
		{\sc askwomen} 		& 0.56 & 0.62 & {\bf 0.65} \\
		{\sc askmen} 		& 0.56 & {\bf 0.66} & {\bf 0.66} \\
		{\sc worldnews} 	& 0.54 & {\bf 0.61} & 0.60 \\
		\hline\rule[-1ex]{0pt}{3.5ex}%
		Improvement 		& - &  12.5\% & 13.2\% \\
		\bottomrule
	\end{tabular}
}
	\caption{Test set ranking NDCG performance for specific subreddits.}
	\label{tab:overallNDCG}
\end{table}

\begin{table}
	\centering
{\footnotesize
	\begin{tabular}{|lccc|}
		\toprule
		{\bf subreddit} 
		& {\bf Pos/Neg}
		& {\bf High/Mid} 
		& {\bf Ranking} 
		\\
		\hline\rule[-1ex]{0pt}{3.5ex}%
		{\sc askscience} & 44.5 & 63.7 & 61.3 \\
		{\sc fitness} 	& 74.7 & 43.9 & 57.5 \\
		{\sc politics} 	& 95.5 & 59.1 & 58.0 \\
		{\sc askwomen} 	& 82.5 & 67.6 & 59.7 \\
		{\sc askmen}	& 87.0 & 66.2 & 60.6 \\
		{\sc worldnews} & 93.3 & 69.9 & 57.3 \\
		\hline\rule[-1ex]{0pt}{3.5ex}%
		Average		& 79.6 & 61.7 & 59.1 \\
		\bottomrule
	\end{tabular}
}
	\caption{Accuracy of binary classifiers trained on balanced data to distinguish: positive vs.\ negative karma (Pos/Neg), high vs.\ mid-level karma (High/Mid), and ranking between any pair (Ranking).}
	\label{tab:diag_classifier}
\end{table}

A major challenge with identifying high karma comments (and negative karma comments) is that they are so rare. Although our feature selection tunes for high rank precision, it is possible that the low-karma data dominate the learning. Alternatively, it may be that language cues are mainly useful for identifying distinguishing the negative or mid-level karma comments, and that the very high karma comments are a matter of timing. To better understand the role of language for these different types, we trained classifiers on balanced data for positive vs. negative karma and high vs.\ mid levels of karma.
For these models, the training pairs could come from different threads, but topic is controlled for in that all topic features are relative (similarity to original post, parent, etc.).
We compared the results to the binary classifier used in ranking, where all pairs are considered. In all three cases, random chance accuracy is 50\%.

Table\,\ref{tab:diag_classifier} shows the pairwise accuracy of these classifiers. We find that distinguishing positive from negative classes is fairly easy, with the notable exception of the more information-oriented subreddit {\sc askscience}. Averaging across the different subreddits, the high vs.\ mid task is slightly easier than the general ranking task, but the variation across subreddits is substantial.  The high vs.\ mid distinction for {\sc fitness} falls below chance (likely overtraining), whereas it seems to be an easier task for the {\sc askwomen,} {\sc askmen}, and {\sc worldnews.}

\section{Related Work}
Interest in social media is rapidly growing in recent years, which includes
work on predicting the popularity of posts, comments and tweets. 
\newcite{danescu2012you} investigate phrase memorability in the movie quotes. \newcite{cheng2014can} explore prediction of information cascades on Facebook. \newcite{weninger2013exploration} analyze the hierarchy of the Reddit discussions, topic shifts, and popularity of the comment, using among the others very simple language analysis.
\newcite{Lampos+14} study the problem of predicting a Twitter user impact score (determined by combining the numbers of user's followers, followees, and listings) using text-based and non-textual features, showing that performance improves when user participation in particular topics is included. 
 
Most relevant to this paper are studies of 
the effect of  language in   popularity predictions.  \newcite{tan2014effect} study how word choice affects the popularity of Twitter messages. As in our work, they control for topic, but they also control for the popularity of the message authors.  On Reddit, we find that celebrity status is less important
than it is on Twitter since on Reddit almost
everyone is anonymous. 
 \newcite{lakkaraju2013s} study how timing and language affect the popularity of posting images on Reddit. They control for content by only making comparisons
between reposts of the same image. Our focus is on studying comments within
a discussion instead of standalone posts, and we analyze a vast majority of language features. \newcite{althoff2014ask} use deeper language analysis on Reddit to predict the success
of receiving a pizza in the Random Acts of Pizza subreddit. 
To our knowledge, this is the first work on ranking comments in terms of community endorsement.

%




\section{Conclusion}

This paper addresses the problem of how language affects the reaction of community in Reddit comments. We collect a new dataset of six subredit discussion forums.  We introduce a new task of ranking comments based on karma in Reddit discussions, which controls for topic and timing of comments. Our results show that using language features improve the comment ranking task in most of the subreddits. Informativeness and relevance are the most broadly useful feature categories; reputation matters for {\sc askscience}, and other categories could either help or hurt depending on the community.
 Future work involves improving the classification algorithm by using new approaches to learning about rare events.

\bibliographystyle{acl}
\bibliography{refs}

\end{document}